\documentclass{egpubl}
\usepackage{egsgp16}

\SpecialIssuePaper         
 \electronicVersion 

\ifpdf \usepackage[pdftex]{graphicx} \pdfcompresslevel=9
\else \usepackage[dvips]{graphicx} \fi

\PrintedOrElectronic

\usepackage{t1enc,dfadobe}
\usepackage{egweblnk}
\usepackage{cite}
\usepackage{amsmath}
\usepackage{algorithm2e}
\usepackage{amssymb}
\usepackage{wrapfig}
\usepackage{pgfplots}
\usepackage{pgf,tikz}
\usepackage{tkz-euclide}
\usepackage{overpic}

\title[Non-Rigid Puzzles]{Non-Rigid Puzzles}

\newcommand{\eg}{\emph{e.g.}}
\newcommand{\ie}{\emph{i.e.}}
\newcommand{\etal}{\emph{et al.}}
\newcommand{\vct}[1]{\ensuremath{\mathbf{#1}}}
\newcommand{\T}{\ensuremath{^\top}}
\newcommand{\C}{\mathbf{C}}
\renewcommand{\L}{\mathbf{L}}
\newcommand{\A}{\mathbf{A}}
\newcommand{\B}{\mathbf{B}}
\newcommand{\uu}{\mathbf{u}}
\newcommand{\vv}{\mathbf{v}}
\newcommand{\G}{\mathbf{G}}
\newcommand{\F}{\mathbf{F}}
\newcommand{\M}{\mathcal{M}}
\newcommand{\N}{\mathcal{N}}

\newcommand{\Nbar}{\overline{\mathcal{N}}}

\newcommand{\orl}[1]{#1}

\newcommand{\gggg}{\mathbf{g}}
\newcommand{\aaa}{\mathbf{a}}

\newcommand{\ones}{\mathbf{1}}

\newcommand{\Tr}{\mathrm{T}}
\author[O. Litany \& E. Rodol\`{a} \& A. M. Bronstein \& M. M. Bronstein \& D. Cremers ]
{O. Litany$^{1,5}$,  
	E. Rodol\`{a}$^{2}$, 
	A. M. Bronstein$^{3,4}$, 
	M. M. Bronstein$^{2,4}$, 
	D. Cremers$^{5}$
	\\
	$^1$Tel Aviv Univeristy, Israel \mbox{   }
	$^2$University of Lugano, Switzerland \mbox{   }
	$^3$Technion, Israel \mbox{   }
	$^4$Intel Perceptual Computing, Israel \mbox{   }
	$^5$TU Munich, Germany
}

\begin{document}
	
 \teaser{
 \vspace{-2ex}
 \includegraphics[width=\linewidth]{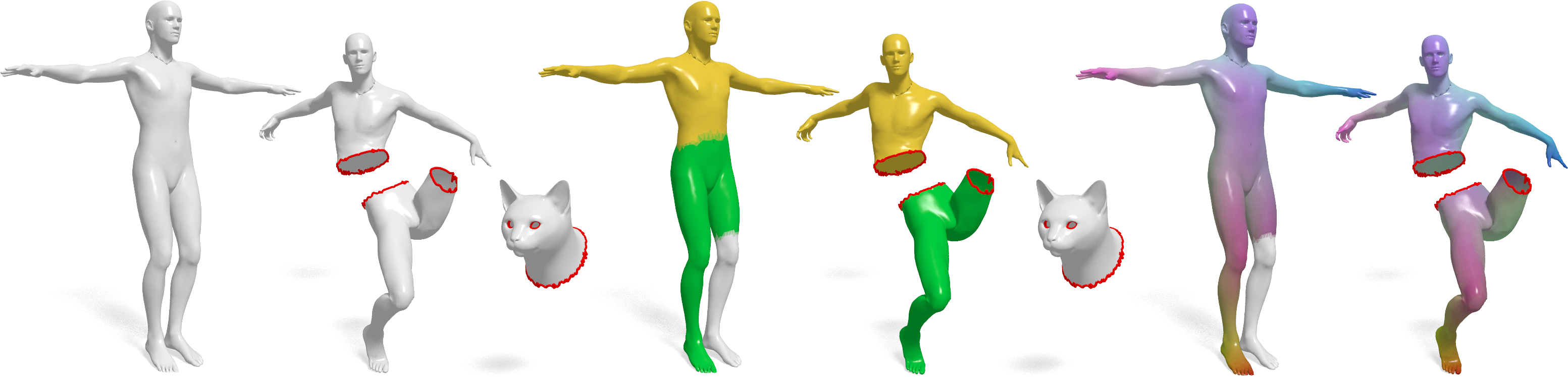}
  \centering
   \caption{ \label{fig:teaser}Example of the non-rigid puzzle problem considered in this paper: given a model human shape (leftmost, first column) and three query shapes (two deformed parts of the human and one unrelated `extra' shape of a cat head), the goal is to find a segmentation of the model shape (second column, shown in yellow and green; white encodes parts without correspondence) into parts corresponding to (subsets of) the query shapes. Third column shows the computed correspondence between the parts (corresponding points are encoded in similar color). }
 }
	
	\maketitle
	
	\begin{abstract}
		Shape correspondence is a fundamental problem in computer graphics and vision, with applications in various problems including animation, texture mapping, robotic vision, medical imaging, archaeology and many more. In settings where the shapes are allowed to undergo non-rigid deformations and only partial views are available, the problem becomes very challenging. To this end, we present a non-rigid multi-part shape matching algorithm. We assume to be given a reference shape and its multiple parts undergoing a non-rigid deformation. Each of these query parts can be additionally contaminated by clutter, may overlap with other parts, and there might be missing parts or redundant ones. Our method simultaneously solves for the segmentation of the reference model, and for a dense correspondence to (subsets of) the parts. Experimental results on synthetic as well as real scans demonstrate the effectiveness of our method in dealing with this challenging matching scenario.

		
	\end{abstract}

\section{Introduction}

Finding correspondence between deformable shapes is one of the cornerstone problems in computer vision and graphics. 
The ability to establish correspondence between 3D geometric data is a crucial ingredient in a broad spectrum of applications ranging  from animation, texture mapping, and robotic vision, to medical imaging and archaeology \cite{van2011survey}. 
The deformable shape correspondence problem comes in a variety of flavors and settings. It is common to distinguish between {\em rigid} and {\em non-rigid} correspondence depending on whether the shapes are allowed to undergo deformations (in this case, one can further distinguish between isometric or inelastic deformations, or more general non-isometric deformations that can also change the shape topology). Second, one distinguishes between {\em full} and {\em partial} correspondence (in the latter case, one allows for some parts of the shapes to be missing; this setting arises in numerous applications that involve real data acquisition by 3D sensors, inevitably leading to missing parts due to occlusions or partial view). Finally, there is also the difference between {\em pairwise} and {\em multiple} correspondence (in the latter case, one tries to establish correspondence between a collection of shapes).
 

\subsection{Related work}

Albeit one of the most broadly studied problems in the domain of geometry processing, correspondence is far from being solved, especially in some challenging settings. We refer the reader to recent survey papers \cite{van2011survey,biasotti2015recent} for an up-to-date review of existing methods. 


\paragraph*{Rigid partial correspondence } problems arising, \eg, in the fusion or completion of multiple 3D scans have been tackled by ICP-like approaches \cite{aiger20084,albarelli-pr15}. 
Bronstein \etal \cite{bronstein2008regularized} used a regularized ICP approach where the matching parts are explicitly modeled, and proposed a functional similar to the Mumford-Shah \cite{mumford1989optimal,vese2002multiphase} imposing part regularity. 
Litany \etal \cite{litany2012putting} extended this approach to multiple rigid shape matching.

\paragraph*{Non-rigid partial correspondence. } 
Several approaches for intrinsic partial matching revolve around the notion of minimum distortion correspondence \cite{bronstein2006generalized}. Bronstein~\etal \cite{bronstein2008not,bronstein2009partial} combined metric distortion minimization with optimization over regular matching parts. Rodol\`{a}~\etal \cite{rodola12,rodola13iccv} relaxed the regularity requirement by allowing sparse correspondences. 
Windheuser \etal \cite{wind11} proposed an integer linear programming solution for dense elastic matching. 
Sahillio{\u{g}}lu and Yemez \cite{sahilliouglu2014partial} proposed a voting-based formulation to match shape extremities, which are assumed to be preserved by the partiality transformation. 
The aforementioned methods are based on intrinsic metric preservation and on the definition of spectral features, hence their accuracy suffers at high levels of partiality -- where the computation of these quantities becomes unreliable due to boundary effects and meshing artifacts.

%

More recent approaches include the alignment of tangent spaces \cite{Brunton201470} and the design of robust descriptors for partial matching \cite{kaick13}. 
Several works tried to employ machine learning methods to deal with partial matches. 
Masci \etal \cite{masci15} introduced Geodesic CNN, a deep learning framework for computing dense correspondences between deformable shapes, providing a generalization of the convolutional networks (CNN) to non-Euclidean manifolds. Wei~\etal \cite{haoli15} focused on matching {\em human} shapes undergoing changes in pose by means of classical CNNs, also tackling partiality transformations.

{\em Dynamic fusion} is a particular setting of the problem, referring to non-rigid tracking of depth images produced by 3D sensors. Attempts to extend ICP-based methods to such a setting \cite{li08} had limited success due to sensitivity to initialization and to the underlying assumption of small deformations. 
Recent works \cite{newcombe2015dynamicfusion,dou20153d} generalizing the Kinect fusion approach \cite{newcombe2011kinectfusion}, were based on volumetric representation of 3D data.

Most of the aforementioned correspondence methods are {\em point-wise}, \ie, one seeks a mapping between vertices of the underlying shapes. Ovsjanikov \etal \cite{ovsjanikov12} introduced {\em functional maps}, representing correspondences between functional spaces on the respective shapes. 
While not intended for partial correspondence, follow-up works \cite{KovnatskyBBV14} showed that functional maps and similar constructions can handle certain settings with missing parts. 
Rodol{\`a} \etal \cite{rodola16-partial} introduced {\em partial functional correspondence}, an extension of \cite{ovsjanikov12} where matched parts are explicitly modeled and regularized in a manner similar to \cite{bronstein2008not,bronstein2009partial}. This method has achieved state-of-the-art performance on the recent SHREC'16 Partial Correspondence benchmark \cite{SHREC2016p}.

\paragraph*{Multiple shape correspondence} in the rigid settings has been addressed in numerous works, including \cite{huang2006reassembling,dualquat,litany2012putting}. In the non-rigid setting, pointwise and functional maps for large shape collections have been explored in  \cite{huang2013consistent,huangn14,sahilliouglu2014multiple,cosmo16}.

\subsection{Main contributions}

In this paper, we are interested in intrinsic, non-rigid, partial, multiple shape correspondence in a setting which we refer to as {\em non-rigid puzzles} (see Figure~\ref{fig:teaser}). \orl{The motivation in mind is to use this formulation as a first step toward automatic reconstruction of the deformable shapes (dynamic fusion), in which one tries to match multiple scans to a near-isometric general model.} We assume to be given a model shape and multiple query shapes, assumed to be parts of (\orl{ near isometrically}) deformed versions of the model shape, possibly with additional clutter. The query shapes may contain overlapping parts, and the model shape might have `missing' regions that do not correspond to any query shape; conversely, there might be `extra' query shapes that have no correspondence to the model shape.

We present a framework for solving 3D non-rigid puzzle problems. We formulate such problems as partial functional correspondences between the query and model shapes, and alternate between optimization on the part-to-whole correspondence and the segmentation of the model. 
Our method can be considered an extension of \cite{rodola16-partial} for the multiple part setting on one hand, and a non-rigid generalization of the rigid puzzles problem treated in \cite{litany2012putting} on the other.

The rest of the paper is organized as follows. 
In Section~\ref{sec:bg} we overview the basic notions in spectral analysis on manifolds and present the partial functional maps framework. 
Section~\ref{sec:problem} formulates the non-rigid puzzle problem and describes the proposed approach, and Section~\ref{sec:implementation} gives the implementation details. 
Section~\ref{sec:results} presents experimental results, where we show how the method copes with some challenging examples. 
Finally, Section~\ref{sec:concl} concludes the paper.

\section{Background}\label{sec:bg}

We model a shape as a two-dimensional Riemannian manifold $\M$ (possibly with boundary $\partial \M$), endowed with the standard measure induced by the volume form. We denote the space of square-integrable functions on the manifold $\M$ by $L^2(\M) = \{ f: \M \rightarrow\mathbb{R} ~|~ \int_{\M}f^2da <\infty \}$, and use the standard $L^2(\M)$ inner product $\langle f, g\rangle_{\M} = \int_{\M}fg da$. 

Our manifolds are equipped with the intrinsic gradient $\nabla_\M$ and Laplace-Beltrami operator $\Delta_{\M}$, generalizing the corresponding notions from Euclidean spaces to manifolds. By analogy to flat spaces, the Laplacian provides us with all necessary tools for extending Fourier analysis to manifolds. In particular, it admits an eigen-decomposition 
 \begin{eqnarray}
\Delta_{\mathcal{M}} \phi_i(x) = \lambda_i \phi_i(x)  & \,\,\,\,\,& x \in \mathrm{int}(\mathcal{M}) \\
\langle \nabla_{\mathcal{M}} \phi_i(x) , \hat{n}(x) \rangle = 0 &\,\,\,\,\,& x \in \partial\mathcal{M}, 
\label{eq:neumann}
\end{eqnarray}
with homogeneous Neumann boundary conditions~(\ref{eq:neumann}), where $\hat{n}$ is the normal vector to the boundary. Here, $0 = \lambda_1 \leq\lambda_2 \leq \hdots$ are eigenvalues and  $\phi_1, \phi_2, \hdots$ are the corresponding eigenfunctions forming an orthonormal basis of $L^2(\M)$.

Since the eigenfunctions of the Laplacian form a basis, any function $f\in L^2(\M)$ can be represented via the (manifold) Fourier series expansion
\begin{eqnarray}\label{eq:fourier}
f(x) &=& \sum_{i\geq 1}  \langle f, \phi_i\rangle_{\M} \phi_i(x)\,.
\end{eqnarray}

\paragraph*{Functional correspondence.}
A recent paradigm shift in the shape matching problem was introduced by Ovsjanikov~\etal \cite{ovsjanikov12}. The authors proposed to model  correspondences among two shapes by means of a linear operator $T: L^2(\M) \rightarrow L^2(\mathcal{N})$, mapping functions on $\M$ to functions on $\mathcal{N}$. Classical point-to-point matching can then be seen as a special case where one maps delta functions to delta functions.

Because $T$ is a linear operator, it can be equivalently represented by a matrix of coefficients $\C=(c_{ij})$ arising from the following short computation: Let us be given orthonormal bases $\{\phi_i\}_{i\geq 1}$ and $\{\psi_i\}_{i\geq 1}$  on $L^2(\M)$ and $L^2(\N)$, respectively, and let us fix some function $f\in L^2(\M)$. Then
\begin{eqnarray}
T f &=& 
T  \sum_{i\geq 1} \langle f, \phi_i \rangle_{\M} \phi_i 
= \sum_{i\geq 1} \langle f, \phi_i \rangle_{\M} T \phi_i \nonumber\\\label{eq:tf}
&=& \sum_{ij\geq 1} \langle f, \phi_i \rangle_{\M} 
\underbrace{\langle T\phi_i, \psi_j \rangle_{\mathcal{N}}}_{c_{ij}} \psi_j\,.
\end{eqnarray}
The application of $T$ is expressed by linearly transforming the expansion coefficients of $f$ from basis $\{\phi_i\}_{i\geq 1}$ onto basis $\{\psi_i\}_{i\geq 1}$.

Choosing as the bases the eigenfunctions $\{\phi_i\}_{i\geq 1}$, $\{\psi_i\}_{i\geq 1}$ of the respective Laplacians on the two shapes yields a particularly convenient representation for the functional map \cite{ovsjanikov12}. By analogy with Fourier analysis, this choice allows to truncate the series \eqref{eq:tf} after the first $k$ coefficients, which is equivalent to taking the upper left $k \times k$ submatrix of $\C$ as an approximation of the full map. Further, one obtains $c_{ij} = \langle T \phi_i,\psi_j\rangle_\mathcal{N} \approx \pm\delta_{ij}$ whenever the two shapes are nearly isometric.  This results in matrix $\C$ being diagonally dominant, since $c_{ij}\approx 0$ if $i\neq j$ . This particular structure was exploited in \cite{pokrass13,kovnatsky13} as a prior for shape matching problems.

\paragraph*{Partial functional correspondence.}
%

\begin{figure}[t]
	\includegraphics[width=\linewidth]{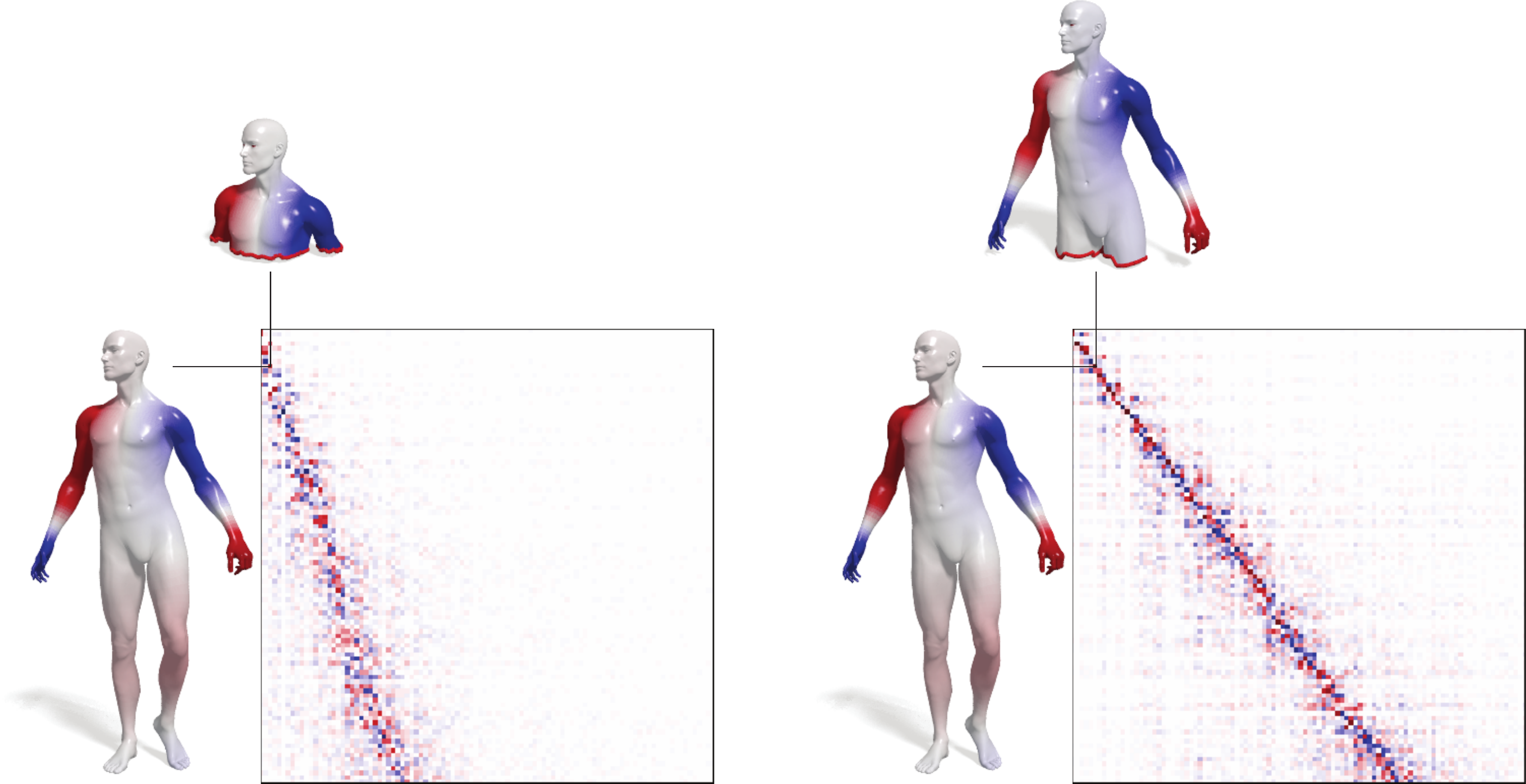}
	\caption{\label{fig:slanted} A key observation behind partial functional maps is the eigenfunction interleaving property, by which the eigenvectors of the full shape contain a subset whose restriction constitutes also the (approximate) eigenvectors of a part of the shape. As a result, the inner products of the eigenvectors of the full shape and the part, restricted to the corresponding subset, form a slanted diagonal matrix. The slope of the slant depends on the ratio of the areas of the partial and the full shape.}
\end{figure}

\begin{figure}[t]
		\begin{overpic}
		[trim=0cm 0cm 0cm 0cm,clip,width=1\linewidth]{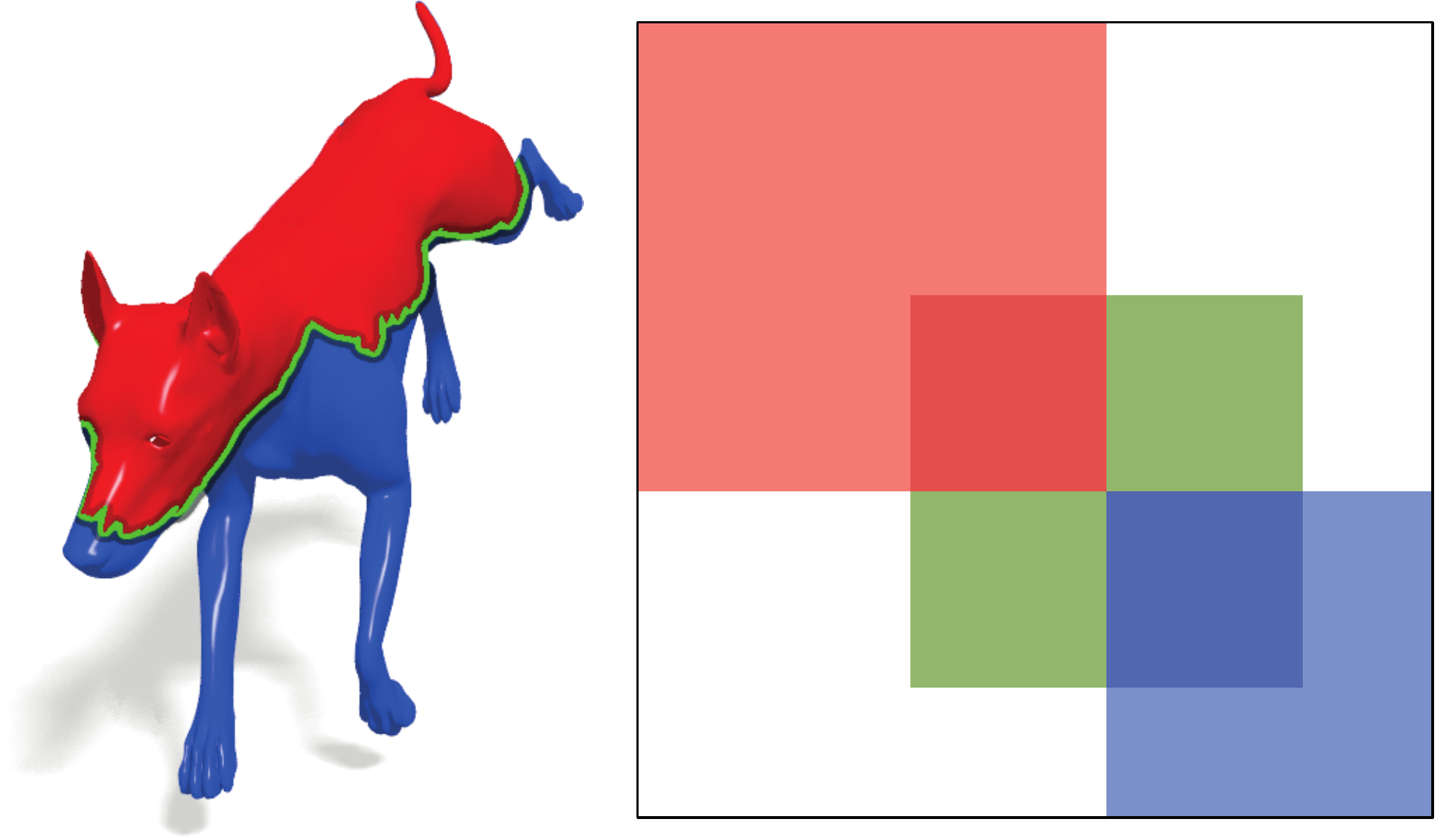}
	\put(52,44){\footnotesize $\mathbf{L}_\N$}
	\put(89.5,5.5){\footnotesize $\mathbf{L}_{\Nbar}$ }
%
%
	\put(80,28.5){\footnotesize $\mathbf{E}$}
	\put(68,16.5){\footnotesize $\mathbf{E}^\top$}
	\put(20,38.5){\footnotesize $\N$}	
	\put(20.25,27){\color{white}\footnotesize $\Nbar$}		
		\end{overpic}
	\caption{\label{fig:slantedb} Structure of Laplacian of a shape consisting of two parts (reproduced from \cite{rodola16-partial}).}
\end{figure}

Let $\mathcal{N}$ be {\em part} of a shape that is nearly isometric to a full model $\M$.
 Recently,  Rodol\`{a} \etal \cite{rodola16-partial}  showed that for each eigenfunction $\psi_j$ of $\mathcal{N}$ there exists a corresponding eigenfunction $\phi_i$ of $\M$ for some $i \ge j$, such that $c_{ij} = \langle T \phi_i,\psi_j\rangle_\mathcal{N} \approx \pm 1$, and zero otherwise. 
Differently from the full-to-full case of \cite{ovsjanikov12}, where approximate equality holds for $i=j$, here the inequality ${i\ge j}$ induces a {\em slanted}-diagonal structure on matrix $\C$ (Figure~\ref{fig:slanted}). In particular, 
 the angle of the diagonal can be precomputed and used as a prior for the matching process.


%

%
The key idea behind their analysis is to model partiality as a perturbation of the Laplacian matrices $\L_\M$, $\L_\N$ of the two shapes. Specifically, consider the dog shape $\M$ shown in Figure~\ref{fig:slantedb}, and assume a vertex ordering where the points contained in the red region $\N$ appear before those of the blue region $\Nbar$. Then, the Laplacian of the full shape $\L_\M$ will assume the structure
\begin{align}\label{eq:blkdiag}
\L_\M = \begin{pmatrix}\L_\N&\mathbf{0}\\\mathbf{0}&\L_{\Nbar}\end{pmatrix} + \begin{pmatrix}\mathbf{P}_\N&\mathbf{E}\\\mathbf{E}\T&\mathbf{P}_{\Nbar}\end{pmatrix}\,,
\end{align}
where the second matrix encodes the perturbation due to the boundary interaction between the two regions. Such a matrix is typically very sparse and low-rank, since it contains non-zero elements only in correspondence of the edges connecting $\partial\N$ to $\partial\Nbar$.

If the perturbation matrix is identically zero, then \eqref{eq:blkdiag} is exactly block-diagonal; this describes the case in which $\N$ and $\Nbar$ are disjoint parts, and the eigenpairs of $\L_\M$ are an interleaved sequence of those of the two blocks. The key result shown in \cite{rodola16-partial} is that this interleaving property still holds even when considering the full matrix $\L_\M$ as given in \eqref{eq:blkdiag}: Its eigenpairs consist of those of the blocks $\L_\N$, $\L_{\Nbar}$, up to some bounded perturbation that depends on the length and position of the boundary $\partial\N$. In other words, the eigenfunctions and eigenvalues of the parts show up among those of the full shape. By letting $\{\phi_i\}_{i\geq 1}$ and $\{\psi_j\}_{j\geq 1}$ denote the eigenfunctions on $\M$ and $\N$ respectively, this is what makes the equality $c_{ij} = \langle T \phi_i,\psi_j\rangle_\mathcal{N} \approx \pm 1$ hold approximately for $i \ge j$.

%
The slant of $\C$ identifying the pairs $(i,j)$ for which $c_{ij}\neq 0$, can easily be computed as follows.
A classical result due to Weyl \cite{weyl11} describes the asymptotic behavior of the Laplacian eigenvalues on manifolds. Let $\lambda_j$ and $\lambda_i$ denote the eigenvalues for shapes $\N$ and $\M$ respectively. Then, Weyl's theorem applied to 2-manifolds states that $\lambda_j \sim \frac{1}{|\N|} j$ and $\lambda_i \sim \frac{1}{|\M|} i$ as $i,j\to\infty$ (here $|\,\cdot\,|$ denotes the surface area). In other words, eigenvalues have a linear growth with the rate inversely proportional to surface area. By the previous analysis, we know that $\lambda_i \approx \lambda_j$ for $i \ge j$. Hence, it follows immediately that matrix $\C$ has a slant given by $\frac{j}{i} \approx \frac{|\N|}{|\M|}$, the area ratio of the two shapes.

The following observation is crucial for the entire paper: Let $M \subset \M$ and $N \subset \N$ be two parts of the shapes $\M$ and $\N$. Then, the functional correspondence between the parts $M$ and $N$ represented in the pair of orthogonal bases $\{\phi_i \}$ and $\{ \psi_j \}$ on $\M$ and $\N$, respectively, has an approximate slanted diagonal structure with the  slant determined by the area ratio $|\N| / |\M|$ (see Figure ~\ref{fig:slanted}). We emphasize that the slanted-diagonal structure of the matrix $\C$ \emph{does not} depend on the parts $M$ and $N$ themselves (which are typically unknown!), but only on the full shapes to which they belong.

\section{Problem formulation}
\label{sec:problem}

\begin{figure}[t]
 \begin{center}
\begin{overpic}[width=\linewidth]{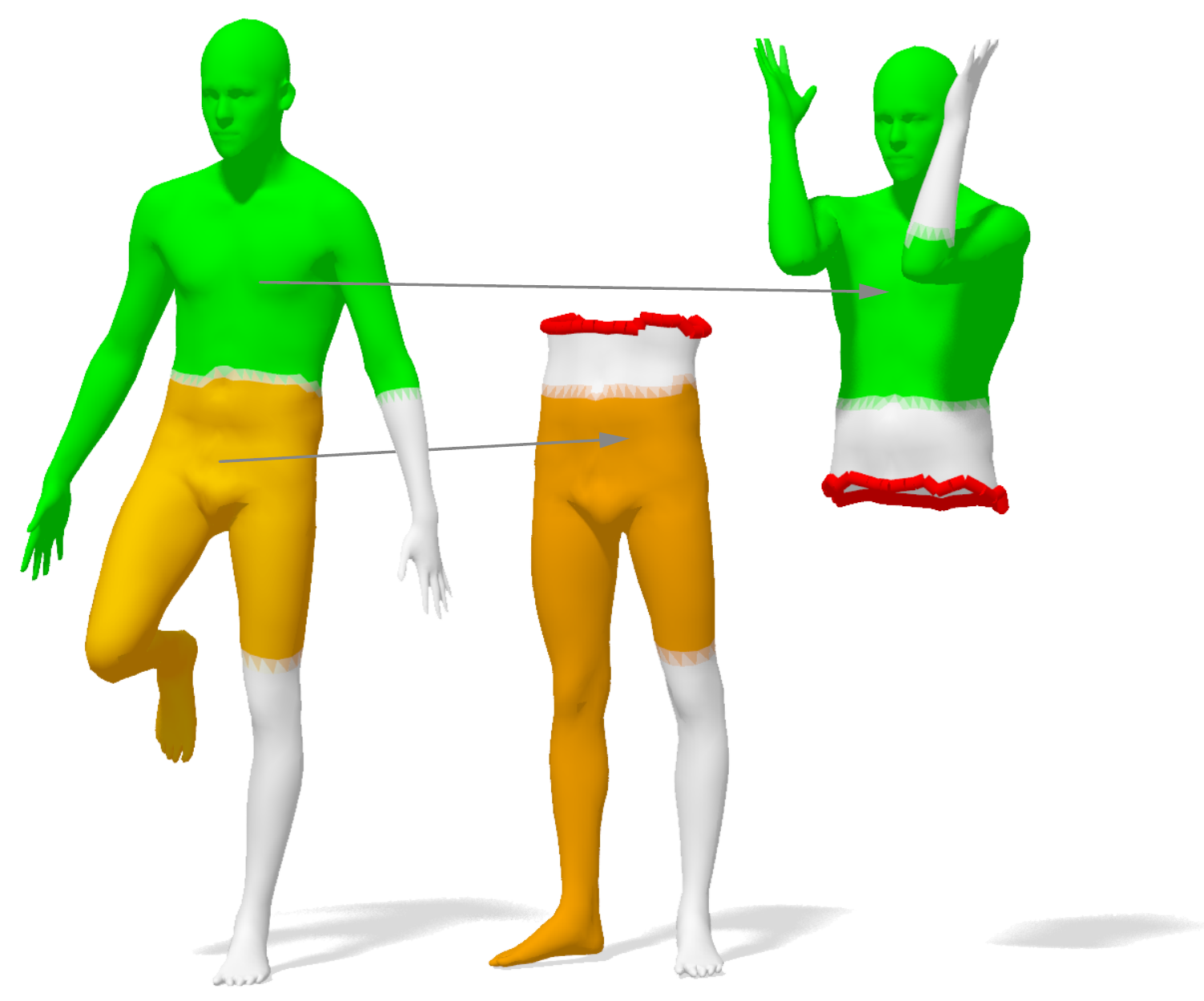}
\put(18,56){$M_1$}
\put(18,46){$M_2$}
\put(50,61){$\tau_1$}
\put(40,48){$\tau_2$}
\put(49,42){$N_2$}
\put(74,54){$N_1$}
\put(25,2){$\M$}
\put(22,20){$M_0$}
\put(62,2){$\N_2$}
\put(85,43){$\N_1$}
\end{overpic}
\end{center}
\caption{\label{fig:notation}The notation we follow in this paper.}
\end{figure}

Let us be given a {\em model shape} $\M$ and a collection $\{ \N_i \}_{i=1}^p$ of $p$ {\em query shapes} constituting possibly incomplete, cluttered, and non-rigidly deformed unknown parts of $\M$. Our goal is to segment $\M$ into $p$ disjoint parts $\{M_i\}$, locate the corresponding parts $\{N_i \subset \N_i \}$ on the input shapes, and calculate the correspondences $\tau_i : M_i \rightarrow N_i$. 
By \emph{clutter} we refer to the regions $N_i^\mathrm{c} = \N_i \setminus N_i$ which are redundant for achieving a full reconstruction. This may include overlaps between the $\N_i$'s, scanning artifacts, and even entire {\em extra parts} coming, \eg, from a different shape as we demonstrate in Figure \ref{fig:teaser}.
By \emph{incompleteness} we mean that the $M_i$'s do not cover $\M$, \ie, there is a {\em missing part}
\begin{equation}
M_0 = \M \setminus \left( \bigcup_{i=1}^p M_i \right)\,.
\end{equation}
$M_0$ can be seen as clutter from the parts perspective.
Figure~\ref{fig:notation} depicts our notation.

We encode the correspondences $\tau_i$ in the functional representation by the matrices $\C_i$ with respect to the Laplacian eigenbasis $\pmb{\Phi}$ of $\M$ (restricted to $M_i$) and the Laplacian eigenbasis $\pmb{\Psi}_i$ of $\N_i$ (restricted to $N_i$).  We further assume to be given as input sets of corresponding functions on each $M_i$ and $N_i$ that are stacked as column vectors of (possibly differently-sized) matrices $\F_i$ and $\G_i$, respectively. 

\orl{\noindent\textbf{Remark~} Since the availability of known corresponding functions is rather a restrictive assumption, in practice we avoid using this input by replacing the $\F_i$'s with a dense descriptor field $\F$ calculated on $\M$ (the number of columns in $\F$ corresponds to the number of dimensions of the descriptor). As the $\G_i$'s, we use the descriptors computed on the corresponding $\N_i$'s. A robust data fitting term accounts for descriptor mismatches.}

With these premises, we formulate the simultaneous segmentation and correspondence as the following optimization problem:
\begin{equation} \label{eq:general_formulation}
\begin{split}
\min_{\C_i, M_i \subseteq \M, N_i \subseteq \N_i} \,\,\, & \sum_{i=1}^p \| \C_i \pmb{\Psi}_i(N_i)^\Tr \G_i - \pmb{\Phi}(M_i)^\Tr \F_i \|_{2,1} \\
&+ \lambda_\M \sum_{i=0}^p R_\mathrm{part}(M_i) + \lambda_\N \sum_{i=1}^p R_\mathrm{part}(N_i) \\
&+ \lambda_\mathrm{corr} \sum_{i=1}^p R_\mathrm{corr}(\C_i) \\
\textrm{s.t.} \quad & M_i \cap M_j = \emptyset \quad \forall i \neq j \\
& M_0 \cup M_1 \cup \dots = \M \\
& |M_i| = | N_i | \geq \alpha |\N_i|,
\end{split}
\end{equation}
where $\pmb{\Phi}(M_i)$ denotes the Laplacian eigenbasis on $\M$ restricted to the part $M_i$, and, similarly, $\pmb{\Psi}_i(N_i)$ denotes the basis on $\N_i$ restricted to the part $N_i$. 

The first term in~(\ref{eq:general_formulation}) is a data fitting term measuring how well the known corresponding functions are mapped between the parts and the model. The $\ell_{2,1}$ norm was chosen here to increase robustness against outliers in the input. This is especially important when one uses descriptors which are not perfectly resilient to non-rigid deformations. 
The second and third terms aggregating $R_\mathrm{part}$ are part regularization terms of the form $R_\mathrm{part}(M) = |\partial M|$ promoting parts with short boundaries and preventing too fragmented segmentation. Note that while the regularization term applies to the missing part $M_0$, the data fitting term does not.  
The last term aggregating $R_\mathrm{corr}$ is a regularization term imposing a prior on the correspondences themselves. Here, the prior comes in the form of a penalty promoting the slanted diagonal structure of each $\C_i$ with the slant proportional to the ratio $|\N_i|/|\M|$ as detailed in the sequel.

Finally, the set of constraints renders the problem a proper segmentation task, enforcing a complete covering and exclusivity of the segments $M_i$. The area constraint enforces the non-cluttered matching areas $N_i$ to be equal. For cases where there exist both clutter in the parts and missing elements, we introduce the inequality term putting a lower bound on the part areas to avoid the trivial solution. In such cases, one has to impose a prior on the resulting non-cluttered area being greater than some percentage $\alpha$ of the entire cluttered part.

Since problem (\ref{eq:general_formulation}) is intractable in its combinatorial formulation, we consider a relaxation of the parts to continuous membership functions $u_i : \M \rightarrow [0,1]$ to encode the $M_i$'s, and $v_i : \N_i \rightarrow [0,1]$ to encode the $N_i$'s.
Assuming that $\M$ is discretized with $m$ vertices, and each $\N_i$ is discretized with $n_i$ vertices, the relaxed and discretized optimization problem can be summarized as
\begin{equation}
\begin{split}
\min_{\C_i, \uu_i, \vv_i} \,
&\sum_{i=1}^p \| \C_i\A_i(\eta(\uu_i)) - \B_i(\eta(\vv_i)) \|_{2,1} + \lambda_\M \sum_{i=0}^p R_\mathrm{part}(\eta(\uu_i)) \\ &+\lambda_\N \sum_{i=1}^p R_\mathrm{part}(\eta(\vv_i)) + \lambda_\mathrm{corr} \sum_{i=1}^p R_\mathrm{corr}(\C_i) \\
\textrm{s.t.} \quad & \sum_{i=0}^{p}\eta({\uu_i}) = 1 \\
&\aaa_{\M}^\Tr \eta(\uu_i) = \aaa_{\N_i}^\Tr \eta(\vv_i) \geq \alpha \aaa_{\N_i}^\Tr \ones  \label{eq:finalprob}
\end{split}
\end{equation}
\orl{Note that, differently from \cite{rodola16-partial}, here we solve $p$ matching problems simultaneously (one per part) under covering and exclusivity constraints.} 

%
\setlength{\columnsep}{5pt}
\setlength{\intextsep}{1pt}
\begin{wrapfigure}{r}{0.365\linewidth}
\begin{center}
%
%
\definecolor{mycolor1}{rgb}{0.00000,0.44700,0.74100}%
\begin{tikzpicture}

\begin{axis}[%
width=0.78\linewidth,
height=0.78\linewidth,
at={(0,0)},
every x tick label/.append style={font=\color{black}, font=\tiny},
every y tick label/.append style={font=\color{black}, font=\tiny},
scale only axis,
xmin=-1,
xmax=2,
ymin=0,
ymax=1,
ytick={0, 1},
axis background/.style={fill=white}
]
\addplot [color=mycolor1,solid,forget plot,line width=1.8pt]
  table[row sep=crcr]{%
-1	1.52299794997646e-08\\
-0.99	1.71717539076965e-08\\
-0.98	1.93610984466908e-08\\
-0.97	2.18295774789823e-08\\
-0.96	2.46127798142481e-08\\
-0.95	2.77508316326447e-08\\
-0.94	3.12889751885592e-08\\
-0.93	3.52782209001035e-08\\
-0.92	3.97760828163563e-08\\
-0.91	4.48474078984518e-08\\
-0.9	5.05653109383886e-08\\
-0.89	5.70122284937469e-08\\
-0.88	6.42811076589922e-08\\
-0.87	7.24767459381326e-08\\
-0.86	8.17173020917217e-08\\
-0.85	9.21359998296012e-08\\
-0.84	1.03883048552245e-07\\
-0.83	1.17127808618722e-07\\
-0.82	1.32061233448777e-07\\
-0.81	1.48898622431837e-07\\
-0.8	1.67882724833035e-07\\
-0.79	1.89287239604763e-07\\
-0.78	2.13420761563388e-07\\
-0.77	2.40631230163935e-07\\
-0.76	2.71310946098335e-07\\
-0.75	3.05902226938048e-07\\
-0.74	3.44903784310624e-07\\
-0.73	3.88877913815033e-07\\
-0.72	4.38458601703662e-07\\
-0.71	4.9436066534847e-07\\
-0.7	5.57390058608664e-07\\
-0.69	6.28455491480384e-07\\
-0.68	7.08581531227992e-07\\
-0.67	7.98923373401816e-07\\
-0.66	9.00783496404767e-07\\
-0.65	1.0156304394715e-06\\
-0.64	1.14511997462152e-06\\
-0.63	1.2911189775644e-06\\
-0.62	1.45573234255902e-06\\
-0.61	1.64133332836158e-06\\
-0.6	1.85059777285668e-06\\
-0.59	2.0865426701433e-06\\
-0.58	2.3525696654092e-06\\
-0.57	2.65251409525824e-06\\
-0.56	2.99070027975912e-06\\
-0.55	3.37200386368863e-06\\
-0.54	3.8019221053065e-06\\
-0.53	4.28665312579568e-06\\
-0.52	4.83318526162124e-06\\
-0.51	5.44939780761089e-06\\
-0.5	6.14417460220729e-06\\
-0.49	6.92753209280417e-06\\
-0.48	7.81076372508016e-06\\
-0.47	8.80660273838707e-06\\
-0.46	9.9294057117616e-06\\
-0.45	1.11953595051117e-05\\
-0.44	1.26227145769708e-05\\
-0.43	1.42320480395197e-05\\
-0.42	1.60465602389026e-05\\
-0.41	1.80924091330859e-05\\
-0.4	2.03990872799098e-05\\
-0.39	2.29998468647685e-05\\
-0.38	2.59321788830813e-05\\
-0.37	2.92383533753715e-05\\
-0.36	3.29660284854039e-05\\
-0.35	3.71689371028716e-05\\
-0.34	4.19076609629032e-05\\
-0.33	4.72505033288217e-05\\
-0.32	5.3274472797904e-05\\
-0.31	6.00663923585598e-05\\
-0.3	6.77241496197523e-05\\
-0.29	7.63581061492546e-05\\
-0.28	8.60926861275813e-05\\
-0.27	9.70681670796059e-05\\
-0.26	0.000109442698320528\\
-0.25	0.000123394575986258\\
-0.24	0.000139124807156554\\
-0.23	0.000156859999666858\\
-0.22	0.000176855618799987\\
-0.21	0.000199399657211052\\
-0.2	0.000224816770233283\\
-0.19	0.000253472935216437\\
-0.18	0.000285780700841898\\
-0.17	0.000322205100531325\\
-0.16	0.000363270313218089\\
-0.15	0.000409567164986024\\
-0.14	0.000461761576524056\\
-0.13	0.00052060407411092\\
-0.12	0.000586940496080823\\
-0.11	0.000661724042557044\\
-0.1	0.000746028833836676\\
-0.09	0.000841065162308863\\
-0.08	0.000948196644353749\\
-0.07	0.00106895950244196\\
-0.0599999999999999	0.00120508423377896\\
-0.0499999999999999	0.00135851995042896\\
-0.04	0.0015314617070033\\
-0.03	0.00172638116573715\\
-0.02	0.00194606098508554\\
-0.01	0.00219363335670875\\
0	0.00247262315663477\\
0.01	0.00278699621904227\\
0.02	0.00314121328482941\\
0.03	0.00354029022094648\\
0.04	0.00398986514899413\\
0.05	0.0044962731609412\\
0.0600000000000001	0.00506662933346624\\
0.0700000000000001	0.00570892077702334\\
0.0800000000000001	0.00643210846691866\\
0.0900000000000001	0.00724623959583143\\
0.1	0.00816257115315988\\
0.11	0.00919370536728809\\
0.12	0.0103537375310918\\
0.13	0.0116584165548331\\
0.14	0.0131253183371028\\
0.15	0.0147740316932731\\
0.16	0.0166263561078816\\
0.17	0.0187065099543546\\
0.18	0.0210413470204683\\
0.19	0.0236605781554612\\
0.2	0.0265969935768658\\
0.21	0.0298866798036362\\
0.22	0.0335692232814825\\
0.23	0.0376878905086059\\
0.24	0.0422897718420338\\
0.25	0.0474258731775668\\
0.26	0.0531511363980637\\
0.27	0.0595243659765015\\
0.28	0.0666080355750907\\
0.29	0.0744679451660281\\
0.3	0.0831726964939224\\
0.31	0.0927929531171571\\
0.32	0.10340045145825\\
0.33	0.11506673204555\\
0.34	0.127861566319081\\
0.35	0.141851064900488\\
0.36	0.157095468885453\\
0.37	0.173646647019005\\
0.38	0.191545348561468\\
0.39	0.210818293477747\\
0.4	0.231475216500983\\
0.41	0.253506016662338\\
0.42	0.27687819487561\\
0.43	0.301534783997461\\
0.44	0.327392982932239\\
0.45	0.354343693774204\\
0.46	0.382252125230751\\
0.47	0.410959565941335\\
0.48	0.440286350732807\\
0.49	0.470035948235428\\
0.5	0.5\\
0.51	0.529964051764572\\
0.52	0.559713649267193\\
0.53	0.589040434058665\\
0.54	0.617747874769249\\
0.55	0.645656306225796\\
0.56	0.672607017067761\\
0.57	0.698465216002539\\
0.58	0.72312180512439\\
0.59	0.746493983337662\\
0.6	0.768524783499017\\
0.61	0.789181706522253\\
0.62	0.808454651438532\\
0.63	0.826353352980995\\
0.64	0.842904531114547\\
0.65	0.858148935099512\\
0.66	0.872138433680919\\
0.67	0.88493326795445\\
0.68	0.89659954854175\\
0.69	0.907207046882843\\
0.7	0.916827303506078\\
0.71	0.925532054833972\\
0.72	0.933391964424909\\
0.73	0.940475634023499\\
0.74	0.946848863601936\\
0.75	0.952574126822433\\
0.76	0.957710228157966\\
0.77	0.962312109491394\\
0.78	0.966430776718518\\
0.79	0.970113320196364\\
0.8	0.973403006423134\\
0.81	0.976339421844539\\
0.82	0.978958652979532\\
0.83	0.981293490045645\\
0.84	0.983373643892118\\
0.85	0.985225968306727\\
0.86	0.986874681662897\\
0.87	0.988341583445167\\
0.88	0.989646262468908\\
0.89	0.990806294632712\\
0.9	0.99183742884684\\
0.91	0.992753760404169\\
0.92	0.993567891533081\\
0.93	0.994291079222977\\
0.94	0.994933370666534\\
0.95	0.995503726839059\\
0.96	0.996010134851006\\
0.97	0.996459709779054\\
0.98	0.996858786715171\\
0.99	0.997213003780958\\
1	0.997527376843365\\
1.01	0.997806366643291\\
1.02	0.998053939014914\\
1.03	0.998273618834263\\
1.04	0.998468538292997\\
1.05	0.998641480049571\\
1.06	0.998794915766221\\
1.07	0.998931040497558\\
1.08	0.999051803355646\\
1.09	0.999158934837691\\
1.1	0.999253971166163\\
1.11	0.999338275957443\\
1.12	0.999413059503919\\
1.13	0.999479395925889\\
1.14	0.999538238423476\\
1.15	0.999590432835014\\
1.16	0.999636729686782\\
1.17	0.999677794899469\\
1.18	0.999714219299158\\
1.19	0.999746527064784\\
1.2	0.999775183229767\\
1.21	0.999800600342789\\
1.22	0.9998231443812\\
1.23	0.999843140000333\\
1.24	0.999860875192844\\
1.25	0.999876605424014\\
1.26	0.999890557301679\\
1.27	0.99990293183292\\
1.28	0.999913907313872\\
1.29	0.999923641893851\\
1.3	0.99993227585038\\
1.31	0.999939933607641\\
1.32	0.999946725527202\\
1.33	0.999952749496671\\
1.34	0.999958092339037\\
1.35	0.999962831062897\\
1.36	0.999967033971515\\
1.37	0.999970761646625\\
1.38	0.999974067821117\\
1.39	0.999977000153135\\
1.4	0.99997960091272\\
1.41	0.999981907590867\\
1.42	0.999983953439761\\
1.43	0.99998576795196\\
1.44	0.999987377285423\\
1.45	0.999988804640495\\
1.46	0.999990070594288\\
1.47	0.999991193397262\\
1.48	0.999992189236275\\
1.49	0.999993072467907\\
1.5	0.999993855825398\\
1.51	0.999994550602192\\
1.52	0.999995166814738\\
1.53	0.999995713346874\\
1.54	0.999996198077895\\
1.55	0.999996627996136\\
1.56	0.99999700929972\\
1.57	0.999997347485905\\
1.58	0.999997647430335\\
1.59	0.99999791345733\\
1.6	0.999998149402227\\
1.61	0.999998358666672\\
1.62	0.999998544267657\\
1.63	0.999998708881022\\
1.64	0.999998854880025\\
1.65	0.99999898436956\\
1.66	0.999999099216504\\
1.67	0.999999201076627\\
1.68	0.999999291418469\\
1.69	0.999999371544509\\
1.7	0.999999442609941\\
1.71	0.999999505639335\\
1.72	0.999999561541398\\
1.73	0.999999611122086\\
1.74	0.999999655096216\\
1.75	0.999999694097773\\
1.76	0.999999728689054\\
1.77	0.99999975936877\\
1.78	0.999999786579238\\
1.79	0.99999981071276\\
1.8	0.999999832117275\\
1.81	0.999999851101378\\
1.82	0.999999867938767\\
1.83	0.999999882872191\\
1.84	0.999999896116951\\
1.85	0.999999907864\\
1.86	0.999999918282698\\
1.87	0.999999927523254\\
1.88	0.999999935718892\\
1.89	0.999999942987772\\
1.9	0.999999949434689\\
1.91	0.999999955152592\\
1.92	0.999999960223917\\
1.93	0.999999964721779\\
1.94	0.999999968711025\\
1.95	0.999999972249168\\
1.96	0.99999997538722\\
1.97	0.999999978170423\\
1.98	0.999999980638902\\
1.99	0.999999982828246\\
2	0.999999984770021\\
};
\end{axis}
\end{tikzpicture}%
\end{center}
\end{wrapfigure}
In the problem above, $\aaa$ denote the vectors of discrete area elements on the corresponding shapes, and $\eta(t) = \frac{1}{2}\tanh\left(6(t-\frac{1}{2})\right)+\frac{1}{2}$ is an element-wise non-linear transformation used to restrict the indicators at each vertex to the range $[0, 1]$ (see inset).
\orl{Function $\eta(t)$ was chosen according to \cite{rodola16-partial}.}
The matrices $\A_i(\eta(\uu_i)) = \pmb{\Phi}^\Tr \mathrm{diag}(\uu_i) \F_i$ and
$\B_i(\eta(\vv_i)) = \pmb{\Psi}_i^\Tr \mathrm{diag}(\vv_i) \G_i$ denote the representation coefficients \orl{of the input descriptor fields} restricted to their respective parts.

As the regularization term of the segments we use a discretized version of the intrinsic Mumford-Shah functional introduced in \cite{bronstein2008not}
\begin{equation}
	R_\mathrm{part}(\eta(\uu)) = \int_{\M}\xi(\eta(\uu))\|\nabla_\M \eta(\uu) \| da \approx \aaa_\M^\Tr \gggg \,,
\end{equation}
where $\xi(t) \approx \delta \left(t-\tfrac{1}{2}\right)$, and
the vector $\gggg$ contains as its elements the values of the discretized intrinsic gradient norm of $\eta(\uu)$ computed on the tangent bundle of $\M$, \orl{weighted element-wise by $\xi(\eta(\uu))$}.

For the regularization of the functional maps $\C_i$, we follow \cite{rodola16-partial},
\begin{eqnarray}
R_\mathrm{corr}(\C) &=& \|\C \odot \mathbf{W}\|_\F^2 
+  \lambda_{1} \sum_{i\neq j}(\C^\top\C)_{ij}^2 \nonumber\\
&+&  \lambda_{2} \sum_{i}((\mathbf{C}^\top\mathbf{C})_{ii} - d_i)^2.
\end{eqnarray}
Here, the first term containing an element-wise product of $\C$ with the funnel-shaped weight matrix $\mathbf{W}$ promotes the slanted-diagonal structure of $\C$. The elements of the weight matrix are given by
\begin{eqnarray}
w_{mn} = e^{-\sigma\sqrt{m^2 + n^2}} \| \frac{\mathbf{n}}{\| \mathbf{n} \|} \times ((m,n)\T - \mathbf{p}) \| \,.
\end{eqnarray}
The slanted diagonal of $\mathbf{W}$ is a line segment $\mathbf{\ell}(t) = \mathbf{p} + t  \frac{\mathbf{n}}{\| \mathbf{n} \|}$ with $t \in \mathbb{R}$, where $\mathbf{p}=(1,1)\T$ is the matrix origin, and $\mathbf{n} = (1, |\N_i|/|\M|)\T$ is the line direction with slope $|\N_i|/|\M|$.
The second factor in $w_{mn}$ is the distance from the slanted diagonal $\mathbf{\ell}$, and $\sigma > 0$ regulates the spread around $\mathbf{\ell}$. In our experiments we set $\sigma = 0.03$.
The second term in $R_\mathrm{corr}(\C)$ promotes orthogonality of $\C$ (\orl{area-preserving maps}), while the third term \orl{regularizes its rank by setting $\{d_i\}_{i=1}^r = 1$ and ${\{d_i\}_{i=r+1}^k = 0}$. Following \cite{rodola16-partial}, the estimation of $r$ is done by 
\begin{eqnarray}
r = \max \{i~|~\lambda_i^\N < \max_j \lambda_j^\M\}\,.
\end{eqnarray}

}

%
%
%
%
%
\begin{figure}[t]
\center
\includegraphics[width=0.85\linewidth]{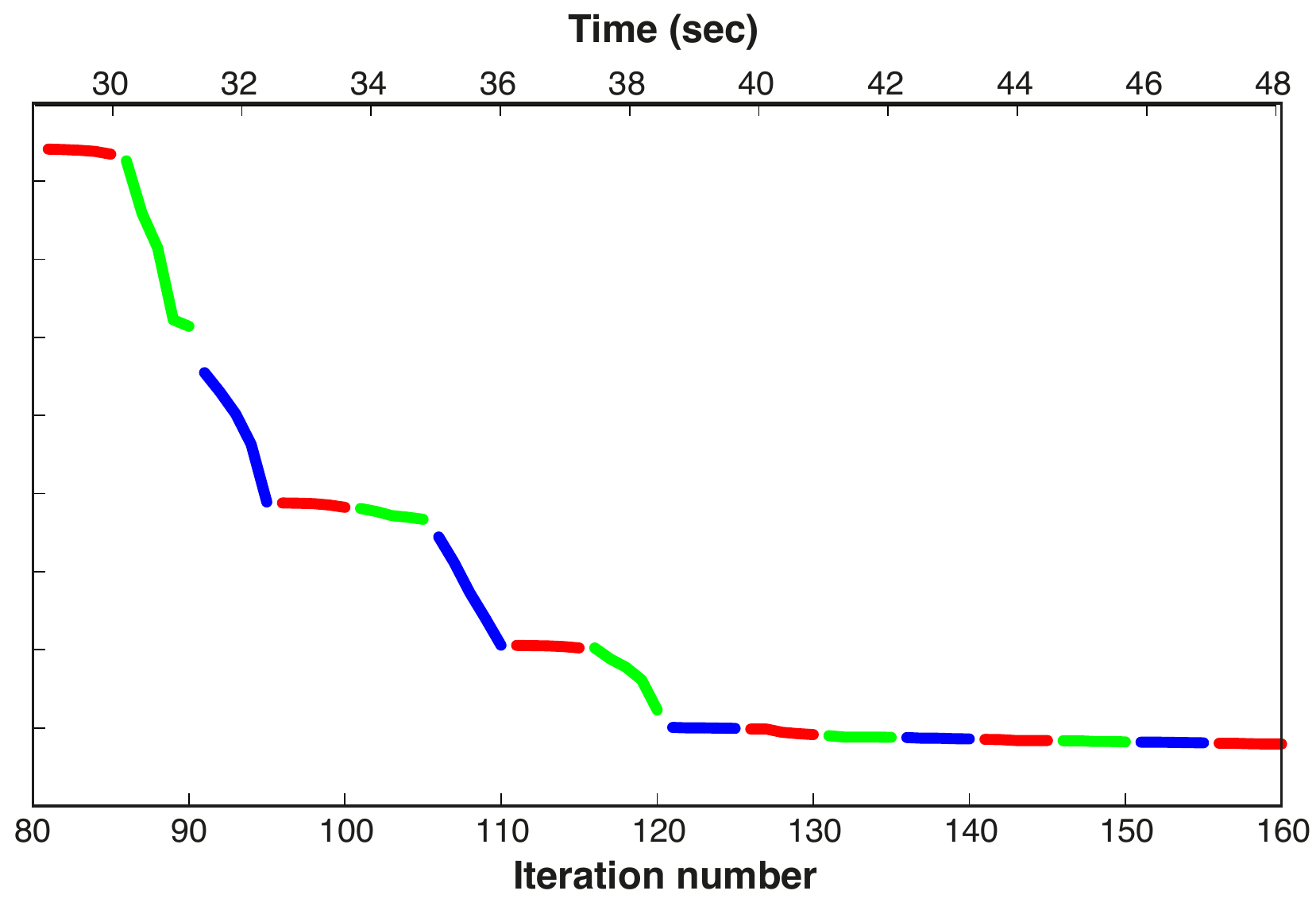}
\caption{\label{fig:iterations} An example showing the decrease in cost during the alternating minimization of the different sets of variables, $\C_i$ (red), $\vct{v}_i$ (green) and $\vct{u}_i$ (blue).}
\end{figure}
%
\section{Implementation}
\label{sec:implementation}
We solve the optimization problem by means of a threefold alternating minimization. To this end, we make use of the masks $\mathbf{W}_i$ to initialize the matrices $\C_i$ by applying the transformation $\C_i = \mathbf{1}\mathbf{1}\T -\frac{\mathbf{W}_i}{\max(\mathbf{W}_i)}$ . We then minimize over the partial functional maps $\C_i$, the model indicator functions $\uu_i$ and the parts indicator functions $\vv_i$ in a cyclic manner, keeping the other parameters fixed (\orl{see Algorithm~\ref{fig:algo}}). 

\begin{figure*}[t]
\centering
\begin{overpic}
		[trim=0cm 0cm 0cm 0cm,clip,width=0.85\linewidth]{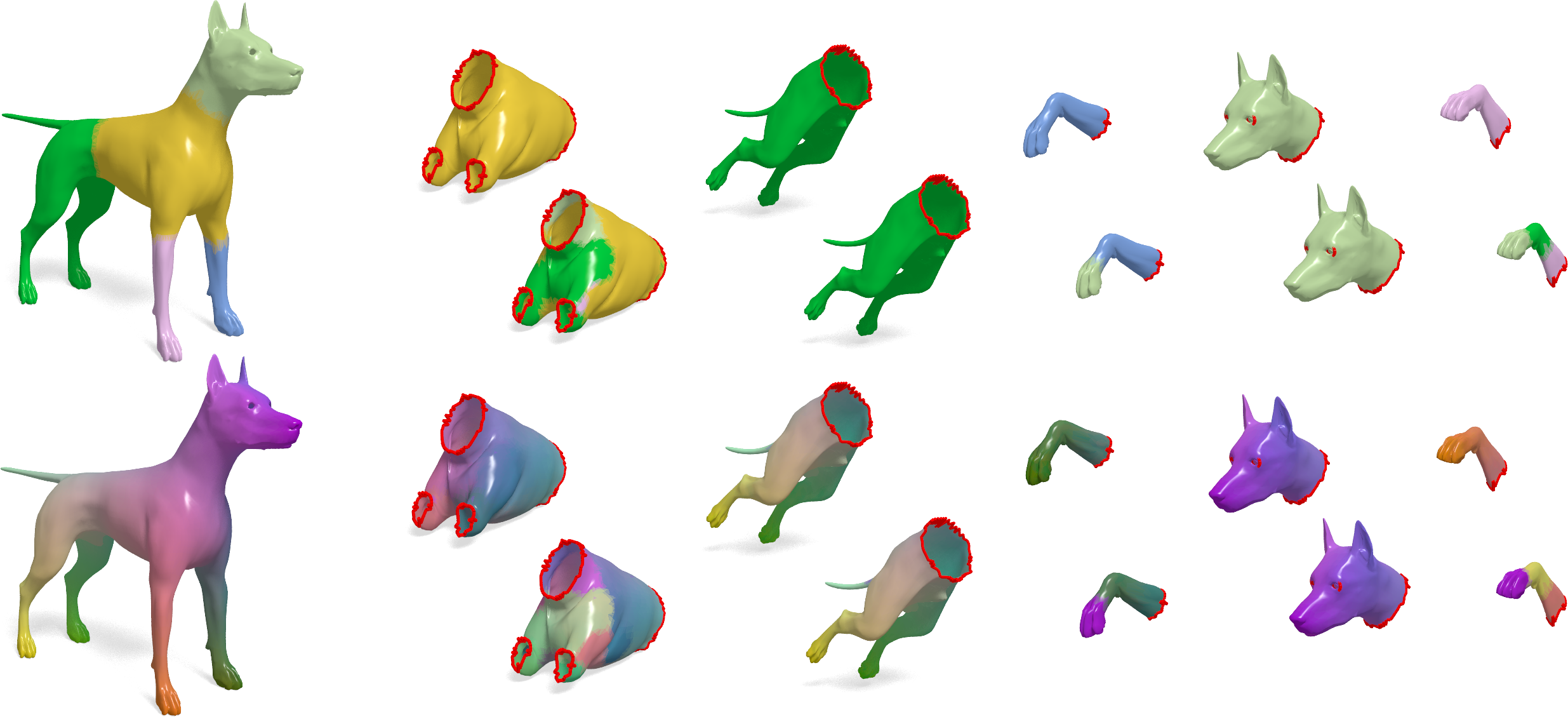}
	\put(20,36){\footnotesize Ours}
	\put(25,27){\footnotesize PFM}
	\put(20,14){\footnotesize Ours}
	\put(25,5){\footnotesize PFM}
		\end{overpic}
	\caption{\label{fig:perfect}Comparison between partial functional maps \cite{rodola16-partial} and our method in a ``perfect'' puzzle setting. For each method we show the membership functions of each part with respect to the model (first two rows), and the color-coded correspondence between parts and model (last two rows). For PFM, the fact that each part is matched independently leads to different parts covering overlapping areas on the model (see, \eg, the four legs). This ambiguity is completely resolved by our method as all parts are matched jointly to the template, yielding a regularizing effect on the correspondence.}
\end{figure*}

\begin{figure}[b]
\centering
\begin{overpic}
		[trim=0cm 0cm 0cm 0cm,clip,width=0.95\linewidth]{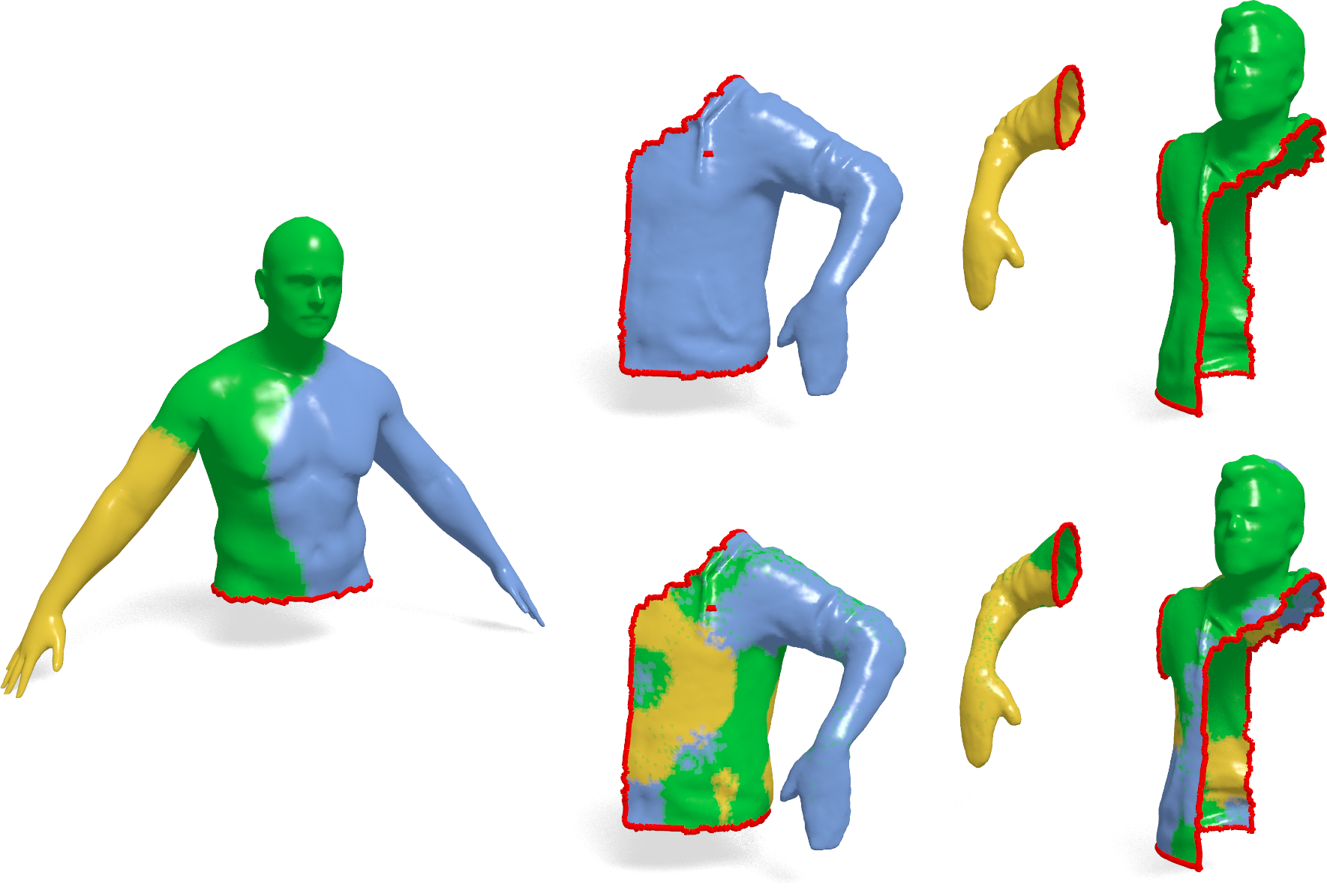}
	\put(16.5,16){\footnotesize template}
		\end{overpic}
\caption{\label{fig:or}Comparison between our method (top row) and PFM (bottom row) on real data. The parts shown on the right were acquired with a 3D scanner.}
\end{figure}

Although this algorithm is not guaranteed to converge, in practice we observed a strictly decreasing cost value as the one shown in Figure ~\ref{fig:iterations}. For the different minimization steps we used the conjugate gradient solver supplied by the Manopt toolbox \cite{manopt}. Since this solver does not support constraints inherently, they were replaced by large quadratic penalties. In order to further refine the solution for the functional maps $\C_i$ we added a $k$-dimensional ICP step \cite{rodola16-partial}. We noticed this step helps especially when the descriptors are performing poorly. The parameters were changed according to the required setting of the experiment. For instance, in the non-isometric experiment (Figure~\ref{fig:or}) we set $\lambda_1 = 0$ to allow changes of areas. 
\begin{algorithm}
	\SetAlgoLined
	\SetKwInOut{Input}{Input}\SetKwInOut{Output}{Output}
	\Input{model $\M$, parts $\{ \N_i \}_{i=1}^p$}
	\Output{segments $\{\uu_i\}$,$\{\vv_i\}$ and maps $\{\C_i\}$}
	initialization: $\C_{i} = \vct{11}\T-\frac{\mathbf{W}_{i}}{\max(\vct{W}_i)}$, 
	$\uu_i = \mathbf{1}$, $\vv_i = \mathbf{1}$ \\
	
	\While{decrease in energy > $\epsilon$ }{
		fix $\{\uu_i\}$ and $\{\vv_i\}$ in Eq.~\eqref{eq:finalprob} and solve for $\{\C_i\}$\;
		run spectral ICP \cite{rodola16-partial} for all $\C_i$\;
		fix $\{\C_i\}$ and $\{\uu_i\}$ in Eq.~\eqref{eq:finalprob} and solve for $\{\vv_i\}$\;
		fix $\{\C_i\}$ and $\{\vv_i\}$ in Eq.~\eqref{eq:finalprob} and solve for $\{\uu_i\}$\;
	}
\caption{\label{fig:algo}\orl{Our pipeline for solving non-rigid puzzles.}}
\end{algorithm}


\section{Experimental results}\label{sec:results}
Our method was implemented in C++/Matlab, and executed on an Intel i7-4710MQ 2.50GHz CPU with 8 logical cores. Typical running times for matching 5 parts to a template of about $10$K vertices were $20$ minutes (end-to-end).
\paragraph*{Data.}
In our experiments we use both synthetic and real data. The synthetic dataset is made up of shapes from the TOSCA \cite{bbk08} and FAUST \cite{bogo14} benchmarks. In order to avoid compatible meshings and make the dataset more realistic, each TOSCA model is independently remeshed to $\sim$10K vertices by iterative pair contractions \cite{garland97}. All FAUST templates are kept at their original resolution ($\sim$7K). The second dataset is composed of real scans acquired with a calibrated Asus Xtion Pro Live RGB-D sensor and then fused into a dense 3D model (about 30K vertices) by DVO-SLAM \cite{kerl13iros}.

The shapes from these datasets are decomposed into a controllable amount of parts by Voronoi decomposition and consensus segmentation\cite{rodola-cgf14}; the former approach leads to generic surface patches having similar area, while the latter tends to produce more semantically meaningful parts (\eg, arms and feet).

\paragraph*{Features.}
Unless differently stated, as dense descriptor fields for the data term in \eqref{eq:general_formulation} we use 350-dimensional SHOT signatures \cite{tombari10}. These are rotation-invariant local features with {\em no} isometry invariance, but whose locality properties result in a higher resilience towards boundary effects than classical spectral features \cite{sun09,aubry11}. Note that we compute dense descriptors for all shape points, including those lying along the boundaries.



 \begin{figure*}[t]
\centering
\begin{overpic}
		[trim=0cm 0cm 0cm 0cm,clip,width=\linewidth]{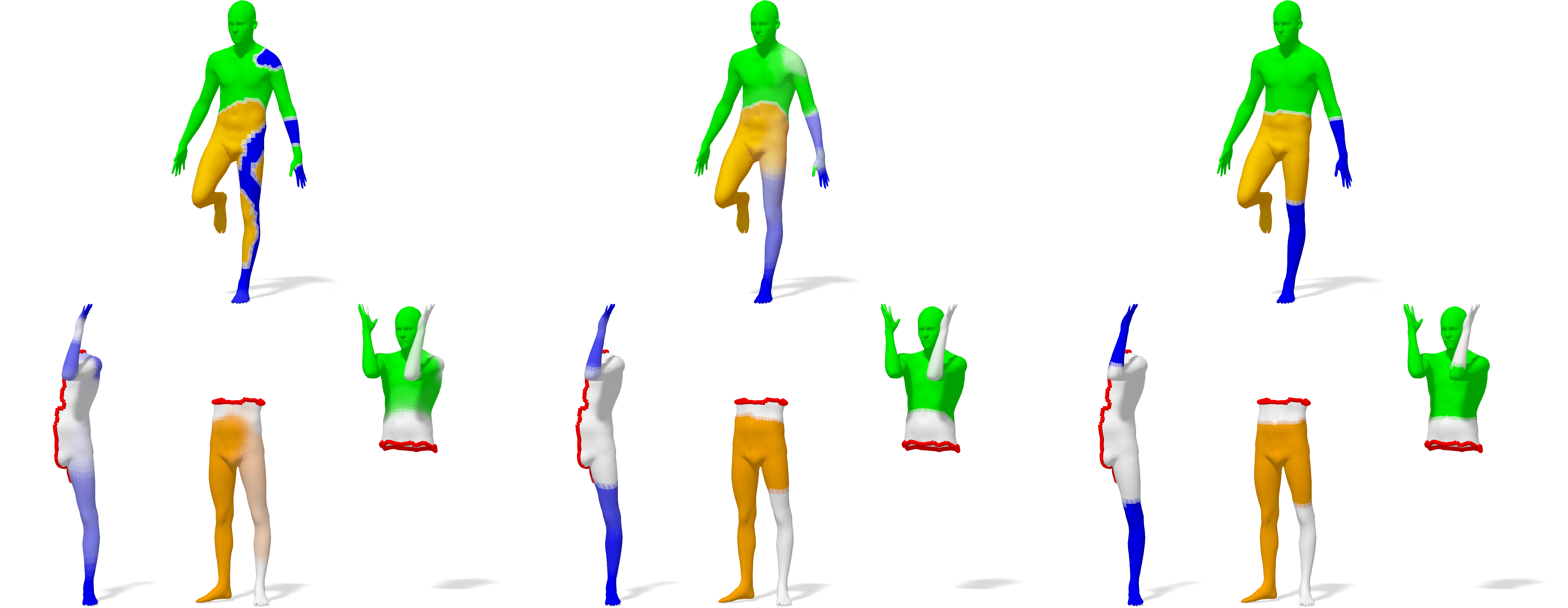}
	\put(20,36){\footnotesize Iteration 1}
	\put(53,36){\footnotesize Iteration 2}
	\put(86,36){\footnotesize Iteration 3}
		\end{overpic}
	\caption{\label{fig:pipeline}An example of our matching pipeline when dealing with overlapping parts. The optimization process alternates between the membership functions on the model (top row) and those on the parts (bottom row). At each alternating step, the membership functions are optimized jointly on the respective parts. Note that in this example there is more than one possible solution (\eg, the blue parts are redundant).}
\end{figure*}

\subsection{Perfect puzzle}
In Figure~\ref{fig:perfect} we show an example of a solution obtained with our method in a basic setting. The input data are five non-overlapping pieces taken from nearly isometric deformations of the model, forming a covering set of the model. SHOT descriptors were used in the data fitting term. No additional clutter is introduced. For this experiment, we compare with the partial functional maps (PFM) method of Rodol\`{a} \etal \shortcite{rodola16-partial} applied to each part separately, resulting in five independent PFM matching problems (one per part).

We performed a similar comparison with real data acquired by a 3D sensor. For this experiment we use the upper part of a shape from FAUST as a template, and portions of a real scanning as the data. Differently from the previous experiment where dense SHOT descriptors are used, here we employ Gaussians supported at $\sim$15 hand-picked matches as data features. The results are reported in Figure~\ref{fig:or}.

\subsection{Overlapping pieces}
\begin{figure*}[t]
\centering
\begin{overpic}
		[trim=0cm 0cm 0cm 0cm,clip,width=\linewidth]{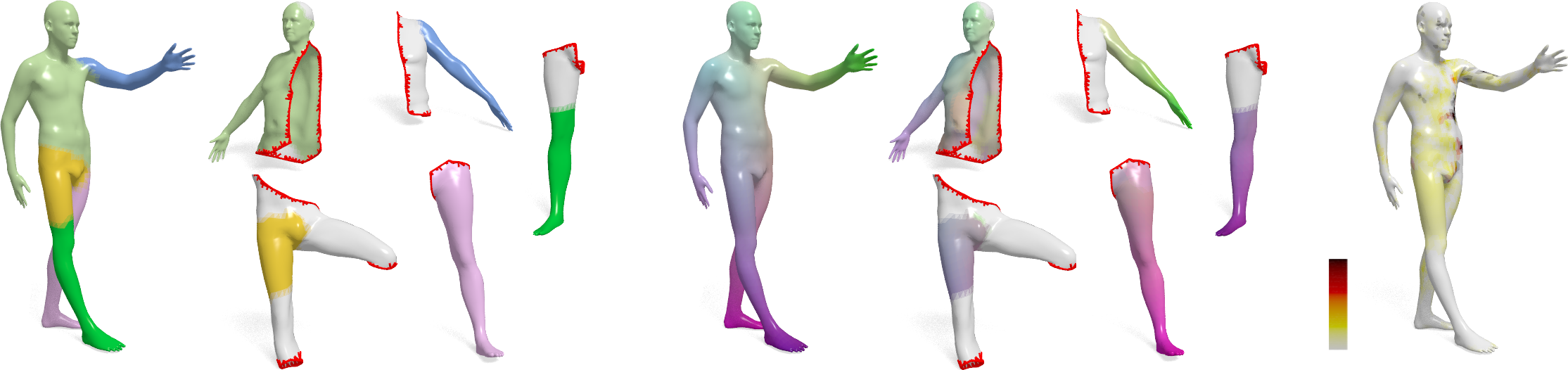}
	\put(86.5,1.5){\footnotesize 0}
	\put(86.5,5.6){\footnotesize 0.1}
		\end{overpic}
	\caption{\label{fig:overlapping}Non-rigid puzzle alignment between overlapping parts. Shown are the final segmentation obtained by our method (left), the dense matchings between the parts and the model (middle), and the normalized geodesic error (shown as a heatmap) to the ground-truth correspondence (right). With the exception of the final column, corresponding points have the same color whereas white color denotes no match. Despite the lack of isometry (two different individuals) and the large overlap, our method correctly identifies non-overlapping subregions on all the parts, providing a perfect covering of the template. Note that this is not the only possible solution, as the optimization problem we solve may have multiple optima.}
\end{figure*}

%
%
A more interesting setup is obtained when allowing the different pieces to have non-zero overlap, as illustrated in Figure~\ref{fig:pipeline}. In Figure~\ref{fig:overlapping} we show additional results obtained in this setting. To make the experiment even more challenging, we produce the input parts by decomposing into five components two different {\em non-isometric} shapes from the FAUST dataset. The decomposition is performed so as to allow large areas of overlap between the pieces. We see that our method copes well with both sources of nuisance even if these show up simultaneously: Overlapping areas are correctly segmented, while the lack of isometry does not have a significant impact on the quality of the correspondence.

\subsection{Incomplete and noisy data}
In practical situations, it may happen that the parts at our disposal do not provide a complete covering of the template model. As described in Section~\ref{sec:problem}, our method naturally allows handling scenarios where some of the parts are missing. This is simply done by introducing a lower bound on the part areas, reflecting some prior knowledge on the amount of missing area; note that, in the absence of clutter, this is directly given by the difference of template area and the sum of the parts.
In practice we implement this by defining a membership function to represent the missing part, which is then treated the same way as the others (\ie, we demand regularity on the missing area, yet provide no data term).
%

\begin{figure}[tbh]
\centering
	\includegraphics[width=\linewidth]{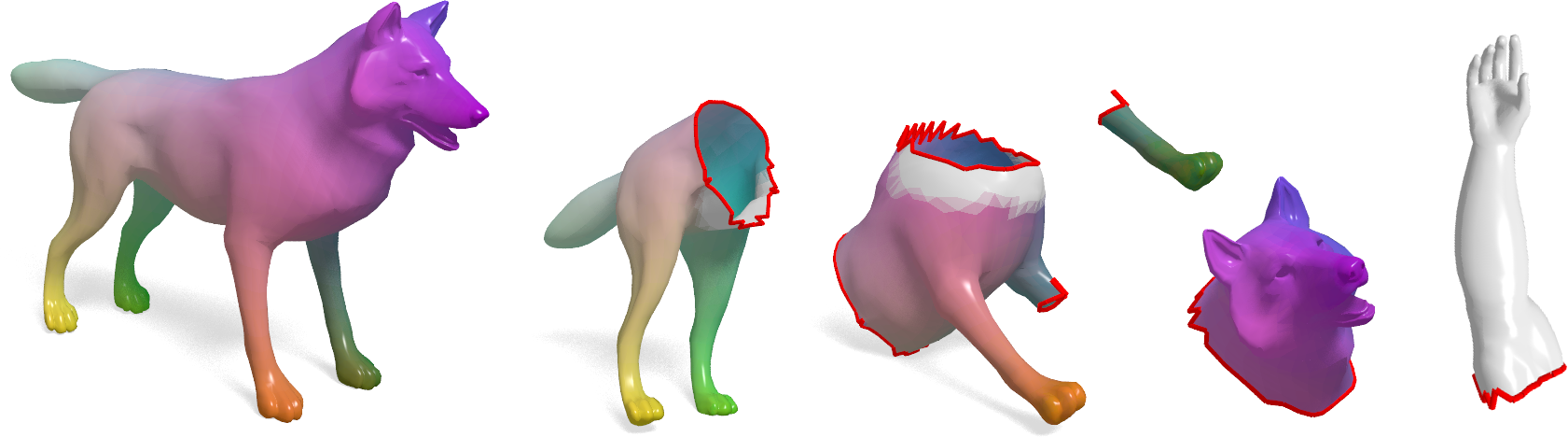}
	\caption{\label{fig:extra}In this example, an additional outlier piece (the human arm) is included in the input set. Our method treats extra pieces as clutter; the arm is automatically selected by the matching process, and completely excluded from the final solution. Note how the presence of the extraneous part did not affect the quality of the correspondence.}
\end{figure}

In Figure \ref{fig:teaser} we show an example of such a scenario, with additional `extra'  pieces that do not belong to the model (the head of the cat). In this noisy setting, the outlier shape is automatically excluded from the final solution due to a lack of mutual support with the rest of the data. Another example of this challenging scenario is given in Figure \ref{fig:extra}.



\section{Discussion and conclusions}\label{sec:concl}
In this paper we introduced a method for solving 3D non-rigid puzzle problems. We formulated the problem as one of partial functional correspondence among an input set of surface pieces and a full template model known in advance. The pieces are matched to the template in a joint fashion, and an optimization process alternates between optimizing for the dense part-to-whole correspondence and the segmentation of the model. We showed how the set of constraints imposed on the plurality of the pieces has a regularizing effect on the solution, leading to accurate part alignment even in challenging scenarios. The framework we presented is remarkably flexible, and can be easily adapted to deal with missing or overlapping pieces, moderate amounts of clutter, and outliers.

\paragraph*{Limitations.}
One of the main limitations, which our method inherits from the functional maps framework, is the need for a reasonably good data term, implying that one has to provide some corresponding functions between the model and the query shapes. 
\orl{This is especially important when the shapes being matched are not nearly isometric.}
In order to allow a fully-automatic pipeline, dense descriptors are used as such corresponding functions. Yet, in real-world settings when the data is contaminated by noise and scanning artifacts, obtaining invariant descriptors is a major challenge. 

\section*{Acknowledgments}
We thank Michael Moeller for the fruitful discussions and Christian Kerl for the technical support in the real scanning experiments.
AB and OL are supported by the ERC Starting Grant No. 335491. ER and MB are supported by the ERC Starting Grant No. 307047. DC is supported by the ERC Consolidator Grant ``3D Reloaded''.


\bibliographystyle{eg-alpha-doi}
\bibliography{egbib}

\newcommand{\etalchar}[1]{$^{#1}$}
\begin{thebibliography}{\uppercase{vKZHCO11}}

\bibitem[AMCO08]{aiger20084}
\textsc{Aiger D., Mitra N.~J., Cohen-Or D.}:
\newblock 4-points congruent sets for robust pairwise surface registration.
\newblock \emph{TOG 27}, 3 (2008), 85.

\bibitem[ART15]{albarelli-pr15}
\textsc{Albarelli A., Rodol\`{a} E., Torsello A.}:
\newblock Fast and accurate surface alignment through an isometry-enforcing
  game.
\newblock \emph{Pattern Recognition 48} (2015), 2209--2226.

\bibitem[ASC11]{aubry11}
\textsc{Aubry M., Schlickewei U., Cremers D.}:
\newblock The wave kernel signature: A quantum mechanical approach to shape
  analysis.
\newblock In \emph{Proc. ICCV Workshops} (2011).

\bibitem[BB08a]{bronstein2008not}
\textsc{Bronstein A.~M., Bronstein M.~M.}:
\newblock Not only size matters: regularized partial matching of nonrigid
  shapes.
\newblock In \emph{Proc. NORDIA} (2008).

\bibitem[BB08b]{bronstein2008regularized}
\textsc{Bronstein A.~M., Bronstein M.~M.}:
\newblock Regularized partial matching of rigid shapes.
\newblock In \emph{Proc. ECCV}. 2008.

\bibitem[BBBK09]{bronstein2009partial}
\textsc{Bronstein A., Bronstein M., Bruckstein A., Kimmel R.}:
\newblock Partial similarity of objects, or how to compare a centaur to a
  horse.
\newblock \emph{IJCV 84}, 2 (2009), 163--183.

\bibitem[BBK06]{bronstein2006generalized}
\textsc{Bronstein A.~M., Bronstein M.~M., Kimmel R.}:
\newblock Generalized multidimensional scaling: a framework for
  isometry-invariant partial surface matching.
\newblock \emph{PNAS 103}, 5 (2006), 1168--1172.

\bibitem[BBK08]{bbk08}
\textsc{Bronstein A., Bronstein M., Kimmel R.}:
\newblock \emph{Numerical Geometry of Non-Rigid Shapes}.
\newblock Springer, 2008.

\bibitem[BCBB15]{biasotti2015recent}
\textsc{Biasotti S., Cerri A., Bronstein A., Bronstein M.}:
\newblock Recent trends, applications, and perspectives in 3d shape similarity
  assessment.
\newblock In \emph{Computer Graphics Forum} (2015).

\bibitem[BMAS14]{manopt}
\textsc{Boumal N., Mishra B., Absil P.-A., Sepulchre R.}:
\newblock {M}anopt, a {M}atlab toolbox for optimization on manifolds.
\newblock \emph{Journal of Machine Learning Research 15} (2014), 1455--1459.
\newblock URL: \url{http://www.manopt.org}.

\bibitem[BRLB14]{bogo14}
\textsc{Bogo F., Romero J., Loper M., Black M.~J.}:
\newblock {FAUST}: Dataset and evaluation for {3D} mesh registration.
\newblock In \emph{Proc. CVPR} (June 2014).

\bibitem[BWW{\etalchar{*}}14]{Brunton201470}
\textsc{Brunton A., Wand M., Wuhrer S., Seidel H.-P., Weinkauf T.}:
\newblock A low-dimensional representation for robust partial isometric
  correspondences computation.
\newblock \emph{Graphical Models 76}, 2 (2014), 70 -- 85.

\bibitem[CRA{\etalchar{*}}16]{cosmo16}
\textsc{Cosmo L., Rodol\`a E., Albarelli A., M\'emoli F., Cremers D.}:
\newblock Consistent partial matching of shape collections via sparse modeling.
\newblock \emph{Computer Graphics Forum} (2016).

\bibitem[CRB{\etalchar{*}}16]{SHREC2016p}
\textsc{Cosmo L., Rodol{\`a} E., Bronstein M.~M., Torsello A., Cremers D.,
  Sahillio\u{g}lu Y.}:
\newblock Shrec'16: Partial matching of deformable shapes.
\newblock In \emph{Proc. 3DOR} (2016).

\bibitem[DTF{\etalchar{*}}15]{dou20153d}
\textsc{Dou M., Taylor J., Fuchs H., Fitzgibbon A., Izadi S.}:
\newblock {3D} scanning deformable objects with a single {RGBD} sensor.
\newblock In \emph{Proc. CVPR} (2015).

\bibitem[GH97]{garland97}
\textsc{Garland M., Heckbert P.~S.}:
\newblock Surface simplification using quadric error metrics.
\newblock In \emph{Proc. SIGGRAPH} (1997), pp.~209--216.

\bibitem[HFG{\etalchar{*}}06]{huang2006reassembling}
\textsc{Huang Q.-X., Fl{\"o}ry S., Gelfand N., Hofer M., Pottmann H.}:
\newblock Reassembling fractured objects by geometric matching.
\newblock \emph{TOG 25}, 3 (2006), 569--578.

\bibitem[HG13]{huang2013consistent}
\textsc{Huang Q.-X., Guibas L.}:
\newblock Consistent shape maps via semidefinite programming.
\newblock In \emph{Computer Graphics Forum} (2013), vol.~32, Wiley Online
  Library, pp.~177--186.

\bibitem[HWG14]{huangn14}
\textsc{Huang Q., Wang F., Guibas L.~J.}:
\newblock Functional map networks for analyzing and exploring large shape
  collections.
\newblock \emph{TOG 33}, 4 (2014), 36.

\bibitem[KBB{\etalchar{*}}13]{kovnatsky13}
\textsc{Kovnatsky A., Bronstein M., Bronstein A., Glashoff K., Kimmel R.}:
\newblock Coupled quasi-harmonic bases.
\newblock \emph{Comput. Graph. Forum 32}, 2pt4 (2013), 439--448.

\bibitem[KBBV14]{KovnatskyBBV14}
\textsc{Kovnatsky A., Bronstein M.~M., Bresson X., Vandergheynst P.}:
\newblock Functional correspondence by matrix completion.
\newblock \emph{CoRR abs/1412.8070} (2014).

\bibitem[KSC13]{kerl13iros}
\textsc{Kerl C., Sturm J., Cremers D.}:
\newblock Dense visual slam for rgb-d cameras.
\newblock In \emph{Proc. IROS} (2013).

\bibitem[LBB12]{litany2012putting}
\textsc{Litany O., Bronstein A.~M., Bronstein M.~M.}:
\newblock Putting the pieces together: Regularized multi-part shape matching.
\newblock In \emph{Proc. NORDIA} (2012).

\bibitem[LSP08]{li08}
\textsc{Li H., Sumner R.~W., Pauly M.}:
\newblock Global correspondence optimization for non-rigid registration of
  depth scans.
\newblock In \emph{Proc. SGP} (2008), pp.~1421--1430.

\bibitem[MBBV15]{masci15}
\textsc{Masci J., Boscaini D., Bronstein M.~M., Vandergheynst P.}:
\newblock Geodesic convolutional neural networks on riemannian manifolds.
\newblock In \emph{Proc. 3dRR} (2015).

\bibitem[MS89]{mumford1989optimal}
\textsc{Mumford D., Shah J.}:
\newblock Optimal approximations by piecewise smooth functions and associated
  variational problems.
\newblock \emph{Comm. Pure and Applied Math. 42}, 5 (1989), 577--685.

\bibitem[NFS15]{newcombe2015dynamicfusion}
\textsc{Newcombe R.~A., Fox D., Seitz S.~M.}:
\newblock {DynamicFusion}: Reconstruction and tracking of non-rigid scenes in
  real-time.
\newblock In \emph{Proc. CVPR} (2015).

\bibitem[NIH{\etalchar{*}}11]{newcombe2011kinectfusion}
\textsc{Newcombe R.~A., Izadi S., Hilliges O., Molyneaux D., Kim D., Davison
  A.~J., Kohi P., Shotton J., Hodges S., Fitzgibbon A.}:
\newblock Kinectfusion: Real-time dense surface mapping and tracking.
\newblock In \emph{Proc. ISMAR} (2011), pp.~127--136.

\bibitem[OBCS{\etalchar{*}}12]{ovsjanikov12}
\textsc{Ovsjanikov M., Ben-Chen M., Solomon J., Butscher A., Guibas L.}:
\newblock Functional maps: a flexible representation of maps between shapes.
\newblock \emph{ACM Trans. Graph. 31}, 4 (July 2012), 30:1--30:11.

\bibitem[PBB{\etalchar{*}}13]{pokrass13}
\textsc{Pokrass J., Bronstein A.~M., Bronstein M.~M., Sprechmann P., Sapiro
  G.}:
\newblock Sparse modeling of intrinsic correspondences.
\newblock \emph{Computer Graphics Forum 32}, 2pt4 (2013), 459--468.

\bibitem[RBA{\etalchar{*}}12]{rodola12}
\textsc{Rodol\`{a} E., Bronstein A., Albarelli A., Bergamasco F., Torsello A.}:
\newblock A game-theoretic approach to deformable shape matching.
\newblock In \emph{Proc. CVPR} (June 2012), pp.~182--189.

\bibitem[RBC14]{rodola-cgf14}
\textsc{Rodol\`{a} E., Bul\`{o} S.~R., Cremers D.}:
\newblock Robust region detection via consensus segmentation of deformable
  shapes.
\newblock \emph{Computer Graphics Forum 33}, 5 (2014), 97--106.

\bibitem[RCB{\etalchar{*}}16]{rodola16-partial}
\textsc{Rodol\`{a} E., Cosmo L., Bronstein M.~M., Torsello A., Cremers D.}:
\newblock Partial functional correspondence.
\newblock \emph{Computer Graphics Forum} (2016).

\bibitem[RTH{\etalchar{*}}13]{rodola13iccv}
\textsc{Rodol\`{a} E., Torsello A., Harada T., Kuniyoshi Y., Cremers D.}:
\newblock Elastic net constraints for shape matching.
\newblock In \emph{Proc. ICCV} (December 2013), pp.~1169--1176.

\bibitem[SOG09]{sun09}
\textsc{Sun J., Ovsjanikov M., Guibas L.}:
\newblock A concise and provably informative multi-scale signature based on
  heat diffusion.
\newblock In \emph{Proc. SGP} (2009).

\bibitem[SY14a]{sahilliouglu2014multiple}
\textsc{Sahillio{\u{g}}lu Y., Yemez Y.}:
\newblock Multiple shape correspondence by dynamic programming.
\newblock In \emph{Computer Graphics Forum} (2014), vol.~33, Wiley Online
  Library, pp.~121--130.

\bibitem[SY14b]{sahilliouglu2014partial}
\textsc{Sahillio{\u{g}}lu Y., Yemez Y.}:
\newblock Partial 3-d correspondence from shape extremities.
\newblock \emph{Computer Graphics Forum 33}, 6 (2014), 63--76.

\bibitem[TRA11]{dualquat}
\textsc{Torsello A., Rodol\`a E., Albarelli A.}:
\newblock Multiview registration via graph diffusion of dual quaternions.
\newblock In \emph{Proc. CVPR} (2011).

\bibitem[TSDS10]{tombari10}
\textsc{Tombari F., Salti S., Di~Stefano L.}:
\newblock Unique signatures of histograms for local surface description.
\newblock In \emph{Proc. ECCV} (2010), pp.~356--369.

\bibitem[VC02]{vese2002multiphase}
\textsc{Vese L.~A., Chan T.~F.}:
\newblock A multiphase level set framework for image segmentation using the
  {Mumford} and {S}hah model.
\newblock \emph{IJCV 50}, 3 (2002), 271--293.

\bibitem[vKZH13]{kaick13}
\textsc{van Kaick O., Zhang H., Hamarneh G.}:
\newblock Bilateral maps for partial matching.
\newblock \emph{Computer Graphics Forum 32}, 6 (2013), 189--200.

\bibitem[vKZHCO11]{van2011survey}
\textsc{van Kaick O., Zhang H., Hamarneh G., Cohen-Or D.}:
\newblock A survey on shape correspondence.
\newblock \emph{Computer Graphics Forum 30}, 6 (2011), 1681--1707.

\bibitem[Wey11]{weyl11}
\textsc{Weyl H.}:
\newblock {\"U}ber die asymptotische {V}erteilung der {E}igenwerte.
\newblock \emph{Nachrichten von der Gesellschaft der Wissenschaften zu
  G{\"o}ttingen, Mathematisch-Physikalische Klasse} (1911), 110--117.

\bibitem[WHC{\etalchar{*}}15]{haoli15}
\textsc{Wei L., Huang Q., Ceylan D., Vouga E., Li H.}:
\newblock Dense human body correspondences using convolutional networks.
\newblock \emph{arXiv 1511.05904} (2015).

\bibitem[WSSC11]{wind11}
\textsc{Windheuser T., Schlickewei U., Schmidt F.~R., Cremers D.}:
\newblock Large-scale integer linear programming for orientation preserving 3d
  shape matching.
\newblock \emph{Computer Graphics Forum 30}, 5 (2011), 1471--1480.

\end{thebibliography}



\end{document}